\documentclass[sigconf]{acmart}

\usepackage{booktabs}
\usepackage{array}
\usepackage{listings}
\usepackage{graphicx}
\usepackage{xspace}

\definecolor{ink}{HTML}{263746}
\definecolor{codeblue}{HTML}{245A92}
\definecolor{codegreen}{HTML}{327451}
\definecolor{codeorange}{HTML}{9B4F16}
\definecolor{codebg}{HTML}{F7F9FC}
\newcommand{\opensac}{\textsc{OpenSaC}\xspace}
\title[\opensac]{\opensac: A Python Library for Search as Code in Deep Research Agents}

\author{Qi Liu}
\email{qiliu6777@gmail.com}
\affiliation{\institution{Renmin University of China}\city{Beijing}\country{China}}

\author{Jiaxin Mao}
\email{maojiaxin@gmail.com}
\affiliation{\institution{Renmin University of China}\city{Beijing}\country{China}}

\author{Fengbin Zhu}
\email{fengbin@nus.edu.sg}
\affiliation{\institution{National University of Singapore}\city{Singapore}\country{Singapore}}

\author{Tat-Seng Chua}
\email{dcscts@nus.edu.sg}
\affiliation{\institution{National University of Singapore}\city{Singapore}\country{Singapore}}

\author{Ji-Rong Wen}
\email{jrwen@ruc.edu.cn}
\affiliation{\institution{Renmin University of China}\city{Beijing}\country{China}}

\begin{document}
\begin{abstract}
Deep research agents investigate complex questions through iterative search,
source examination, and evidence synthesis. A conventional workflow alternates
reasoning with individual search and reading actions. Search as Code introduces
a new paradigm: agents write programs that compose search, source reading, reranking,
model calls, and local processing, returning selected observations for further reasoning.
To support this workflow in existing coding agents, we present \opensac,
an open-source Python library with a companion Search as Code skill. Configurable capability interfaces let agents
combine these operations while retaining evidence
for inspection and reuse. Compared with standard Codex using its built-in web search,
the \opensac configuration uses less execution time and fewer model tokens on
BrowseComp and WideSearch, with higher accuracy on BrowseComp and slightly lower
row F1 on WideSearch.
We aim to lower the barrier to adopting Search as Code and provide an open
foundation for community development and transparent research on deep research agents.
\end{abstract}

\maketitle

\noindent\textbf{Source code:} \url{https://github.com/liuqi6777/OpenSaC}

\begin{figure}[!t]
  \centering
  \includegraphics[width=0.94\columnwidth]{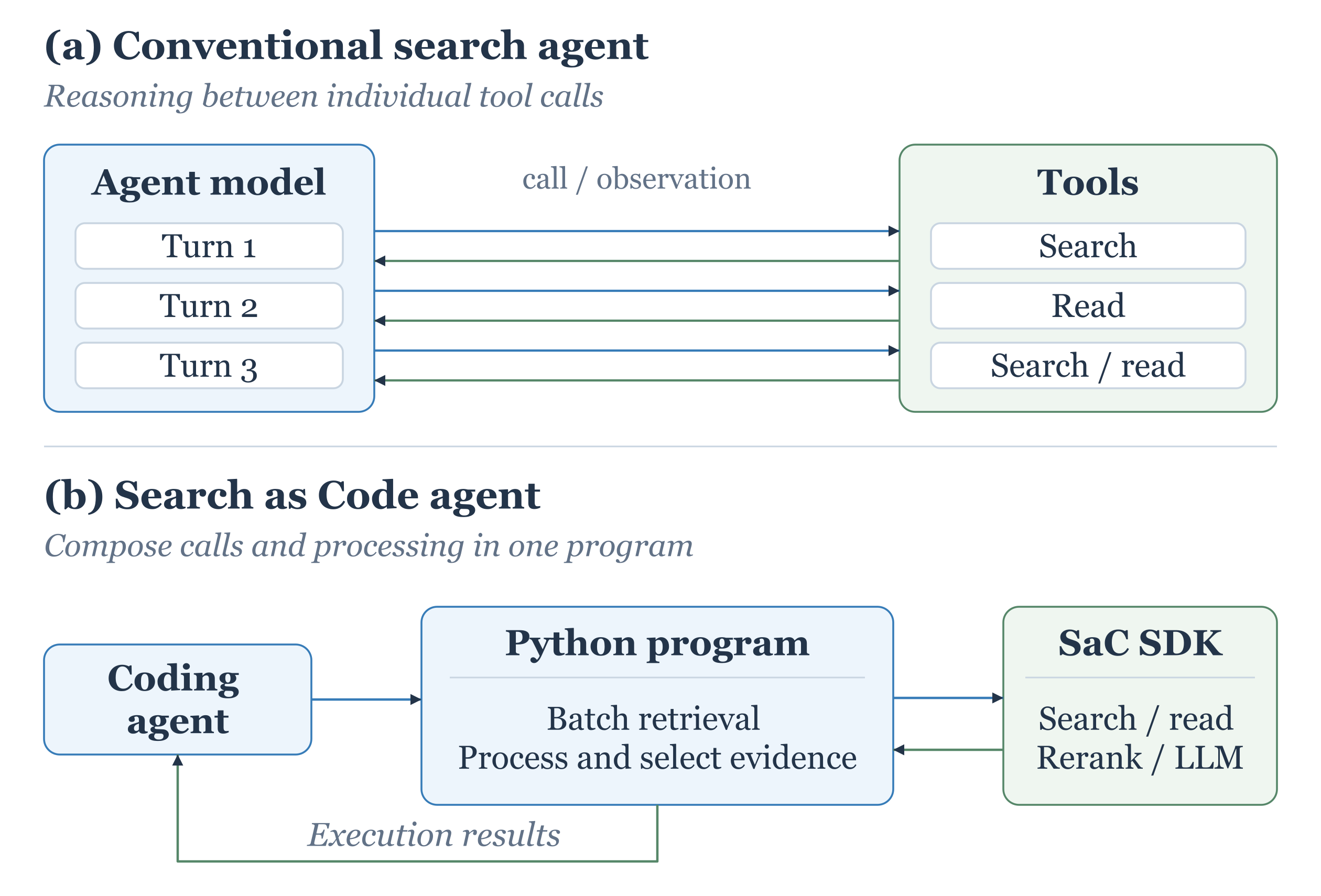}
  \caption{Comparing with tool-calling agent,
  Search as Code composes SDK calls and local processing within a program,
  returning execution results for further reasoning.}
  \Description{The upper panel shows three rounds of calls and observations
  between an agent model and tools. The lower panel shows a coding
  agent writing a Python program that combines SaC SDK calls and local
  processing and returns execution results to the agent.}
  \label{fig:agent-comparison}
\end{figure}

\begin{figure*}[t]
  \centering
  \includegraphics[width=\textwidth]{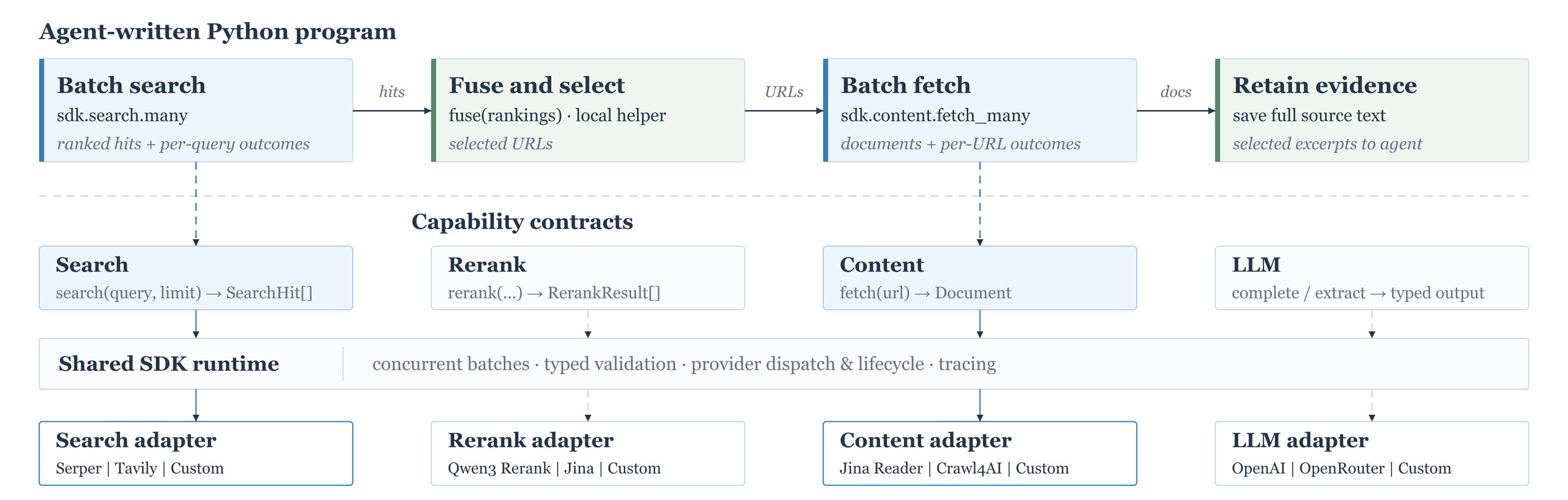}
  \caption{Program composition and provider dispatch in \opensac. \texttt{fuse}
  runs locally, while provider-backed capabilities use a shared runtime for
  concurrency, dispatch, lifecycle management, and tracing. Adapter choices
  illustrate extensibility:
  Tavily, Qwen3 Rerank, Jina Rerank, and Crawl4AI require extensions, while
  OpenRouter is an OpenAI-compatible endpoint. Custom adapters use Python entry
  points. The program is schematic; Listings~\ref{lst:search}
  and~\ref{lst:fetch} show recorded case-study programs (Section~\ref{sec:case-study}).}
  \Description{A program searches, fuses ranked hits, fetches documents, and
  retains full sources while returning selected excerpts. Search and Content
  connect to capability contracts, a shared runtime, and configurable adapters.
  Search, Rerank, Content, and LLM each show named backend choices and a custom
  adapter option. The shared runtime handles concurrent batches, validation,
  provider dispatch and lifecycle management, and tracing; the program does not
  call Rerank or LLM.}
  \label{fig:framework}
\end{figure*}

\newpage
\section{Introduction}
Deep research agents investigate complex questions through iterative information
gathering, source examination, and evidence
synthesis~\cite{openai2025deepresearch,anthropic2025research}. An agent must
break a question into information needs, find relevant sources, reconcile their
contents, and decide whether the evidence is sufficient to answer. For example,
comparing labor markets across states requires locating unemployment, employment,
and earnings series, checking their definitions, and aligning observations across
places and months. Producing a verifiable answer therefore requires coordinating
retrieval with the processing and organization of evidence.

To carry out this process, a conventional approach alternates model reasoning
with individual search and page-reading actions, following the reasoning--action
loop exemplified by ReAct~\cite{yao2023react}. The agent receives tool outputs and
then chooses its next action. This interaction is useful when new evidence
determines what to do next. Within a research task, however, many requests have
inputs that are already known, and operations such as merging candidates,
filtering passages, or parsing records can proceed without another model decision.
Returning every intermediate result to the model adds interactions and context
that these operations do not require. Grouping them into executable programs
allows the agent to reserve further reasoning for decisions that depend on new evidence.

Search as Code, proposed by Perplexity~\cite{perplexity2026sac}, realizes this idea
by making an executable program the unit of interaction. The agent writes code
that combines search, reading, reranking, and model calls with local processing,
then returns selected execution results for further reasoning. Independent searches
can run in a batch,
candidates can be merged before reading, and full sources can be saved while only
relevant passages enter the agent's context. The agent then uses these observations
to decide whether to answer or revise its research plan. The Programmatic Search
Agents (PSA) framework likewise explores executable search cells and selective
evidence presentation~\cite{qian2026psa}. Direct Corpus Interaction (DCI) gives agents
terminal-based access to a locally available corpus~\cite{li2026dci}, whereas
open-web research also requires source discovery and fetching.

Putting this paradigm into practice requires more than access to a hosted search
service or a Python interpreter. Perplexity's public Search
SDK~\cite{perplexity2026sdk} provides hosted search and page snippets through a
Python wrapper around a compiled core; it does not expose the implementation of
the full SaC architecture. Building an open framework for existing coding agents
requires composable interfaces across independently configurable providers,
and agents need guidance on composing calls, retaining sources, and selecting
observations. For transparent research, users must also be able to inspect the
generated program and its evidence, and change a strategy or provider without
rebuilding the surrounding workflow. Our goal is to bring these capabilities into existing coding agents through
an open library and explicit guidance for program composition.

To this end, we present \opensac, an open-source Python library and
companion \emph{Search as Code skill} that provides instructions for composing
research programs in agents such as Codex and Claude Code.
The library provides composable interfaces for search, content fetching,
reranking, and optional language-model calls through independently configurable
providers. The skill guides agents to combine these operations with ordinary
Python in their existing workspace. Together, they let agents batch requests,
process results, retain sources, and return selected evidence within the host
agent's workflow. Figure~\ref{fig:agent-comparison} contrasts the interaction
patterns, and Figure~\ref{fig:framework} outlines the supporting architecture.

We evaluate the \opensac configuration against standard Codex with built-in web search
on BrowseComp~\cite{wei2025browsecomp} and WideSearch~\cite{wong2025widesearch}.
The \opensac configuration uses less execution time and fewer model tokens on both
benchmarks; answer quality improves on BrowseComp and declines slightly on
WideSearch. We also demonstrate with some cases to show how the agent works with
the search-as-Code workflow.
By making generated programs and retained evidence
available for inspection and modification, \opensac aims to lower the barrier to
adopting Search as Code and provide a shared foundation for community development
and transparent academic research on deep research agents.

\section{System Overview}

\subsection{From an agent decision to an observation}
\opensac separates the agent's research decisions from program execution.
The coding agent chooses the next steps and writes a Python program that combines
software development kit (SDK) calls with local computation and file operations.
The library supplies typed
interfaces for search, content fetching, reranking, and model calls; the host
executes the program. Selected execution results return to the agent, which
decides whether to answer or investigate further. These observations can include
source excerpts, extracted records, computed summaries, or errors.

Figure~\ref{fig:framework} illustrates a possible labor-statistics program.
Independent searches cover unemployment, employment, and earnings series;
the program merges rankings, selects URLs, fetches data, and retains responses
before returning records. Grouping known requests leaves the agent to identify
gaps and decide on follow-up queries.

\subsection{Composable capabilities}
Table~\ref{tab:sdk-methods} summarizes the core SDK methods. We expose information
acquisition and evidence processing as separate capabilities so that agents can
compose a workflow around the current information need. A program may stop after
search, fetch known URLs directly, or use reranking and model-based extraction
only when needed. Each capability is configured independently, including model
calls that use an endpoint separate from the coding agent's model.

\begin{table}[!t]
  \centering
  \small
  \setlength{\tabcolsep}{3pt}
  \renewcommand{\arraystretch}{1.12}
  \caption{Core SDK methods. Batch variants preserve input order and return
  per-input data or errors. Local helpers make no provider calls.}
  \label{tab:sdk-methods}
  \begin{tabular*}{\columnwidth}{@{\extracolsep{\fill}}ll@{}}
    \toprule
    Method & Operation \\
    \midrule
    \texttt{search} & Find and rank web pages \\
    \texttt{search.many} & Run independent searches in a batch \\
    \texttt{content.fetch} & Fetch text from a URL \\
    \texttt{content.fetch\_many} & Fetch text from multiple URLs \\
    \texttt{rerank} & Reorder items by query relevance \\
    \texttt{llm.complete} & Generate text from a prompt \\
    \texttt{llm.complete\_many} & Generate text for each prompt \\
    \texttt{llm.extract} & Extract fields matching a schema \\
    \texttt{llm.extract\_many} & Extract fields from multiple texts \\
    \texttt{fuse} & Merge ranked lists into one ranking \\
    \texttt{dedup} & Remove duplicates by URL or key \\
    \bottomrule
  \end{tabular*}
\end{table}

Shared result types let programs connect capabilities without parsing each
backend's response format. Search hits carry URLs, titles, snippets, domains,
and optional dates; fetched documents preserve their requested URLs. Reranking
returns original objects, retaining source identifiers and attached fields.
These choices preserve the connection between processed evidence and its source.
We also distinguish reformulations of one intent, which share a fused ranking,
from independent information needs, whose batch outcomes remain separate.
The agent can thus choose where to combine evidence rather than inherit a fixed pipeline.

\subsection{Program execution and SDK runtime}
The host executes programs and provides a workspace for retained evidence.
The synchronous SDK bridges to an asynchronous runtime that manages capability calls.

\textbf{Supporting Search as Code.} Programs use the host's workspace to retain
sources and intermediate results across executions. They can save full documents
with their URLs, return selected excerpts, and reuse saved text in later programs.
Local helpers complement this workspace: \texttt{fuse} merges ranked lists through
weighted reciprocal rank fusion, and \texttt{dedup} removes repeated identities.
Both preserve original objects, default to exact URL keys, and accept custom keys.
Together, local helpers and ordinary file operations let programs combine
evidence, retain sources, and select what to return to the agent.

\textbf{Supporting reliable execution.} The runtime bounds concurrent provider
requests, applies per-request timeouts, and validates inputs and outputs.
Input-aligned batch outcomes preserve successful results when individual requests
return errors, so programs can inspect failures and selectively retry.
Providers are loaded on first use, reused across calls, and closed with the SDK.
Optional host-controlled logging records capability calls, outcomes, and durations
for inspection. The host controls program execution and the overall task budget.

\subsection{Provider dispatch}
Research workflows may require different search services, content readers, or
model endpoints. Provider dispatch separates these backend choices from program
composition, allowing each capability to be configured independently.

Adapters map shared capability contracts to each backend's request and response
formats. They register through Python entry points; the runtime
routes calls to the configured adapter for each capability. Included adapters
support Serper and Jina Reader, HTTP backends, and model endpoints compatible
with the OpenAI API.

Users can combine services, replace backends while retaining program logic, or
compare provider configurations with the same program. Developers can connect
new services through custom adapters without modifying the core library
(Figure~\ref{fig:framework}).

\subsection{Agent integration}
\opensac pairs the library with a companion Search as Code skill that guides
agents in composing research programs. The library supplies executable
capabilities; the skill explains when to batch independent requests, preserve
sources, select observations, and reuse evidence. This guidance is provided
through the skill mechanisms of coding agents such as Codex and Claude
Code~\cite{openai2026skills,anthropic2026skills}.

The skill works with the agent's existing Python execution and file tools.
Users can revise its guidance to explore different batching, evidence-selection,
and reuse strategies within the same host workflow.

\section{Experiments}
We evaluate answer quality and execution efficiency on two benchmarks, then
illustrate the workflow with a recorded case study.

\subsection{Experimental setup}
We evaluate English BrowseComp fact finding~\cite{wei2025browsecomp} and
WideSearch information collection~\cite{wong2025widesearch} on paired task sets
drawn from 200 tasks per benchmark. We ran both conditions through
Harbor~\cite{harbor2026framework} and archived trajectories, outputs, and grader
results on Harbor Hub; we report paired, audited subsets.
The native condition uses the standard Codex environment with its built-in
web search. The \opensac
condition uses the package and skill, with native web search disabled. Both conditions
use Codex 0.147.0 with \texttt{gpt-5.6-luna}, \texttt{xhigh} reasoning, and a
30-minute task budget.

Quality is answer accuracy for BrowseComp and macro row F1 for WideSearch,
with failed runs included. We exclude 31 BrowseComp tasks and 33 WideSearch tasks
whose saved trajectories show that benchmark answer materials exposed the current
task's answers to the agent in either condition. We further
exclude one WideSearch task with a blocked output stream under \opensac,
leaving 169 and 166 paired tasks, respectively. Reported WideSearch scores include targeted manual corrections for four tasks
in both conditions, addressing grading issues involving units, formatting, and
entity aliases without changing predictions or reference answers.
We report mean agent execution time, cumulative input and output
tokens, and recorded model cost over one run per task. Time excludes setup and
judging; cost excludes retrieval-provider charges.

\begin{table}[t]
  \centering
  \small
  \setlength{\tabcolsep}{3pt}
  \caption{Native retrieval and \opensac on BrowseComp and WideSearch. 
  Quality is accuracy or macro row F1, respectively;
  time, tokens, and recorded model cost are per-task means.
  Failed runs and cached input tokens are included.}
  \label{tab:results}
  \begin{tabular}{@{}llrrrr@{}}
    \toprule
    Benchmark & Condition & Quality (\%) & Time (s) & Tokens (k) & Cost (\$) \\
    \midrule
    BrowseComp & Native & 82.25 & 473.01 & 4102.9 & 0.14324 \\
               & \opensac & 85.21 & 329.59 & 2552.1 & 0.09117 \\
    \midrule
    WideSearch & Native & 60.62 & 767.01 & 5416.0 & 0.18631 \\
               & \opensac & 59.78 & 540.36 & 4667.0 & 0.16286 \\
    \bottomrule
  \end{tabular}
\end{table}

\subsection{Results}
On BrowseComp, \opensac
improves accuracy by 2.96 percentage points while reducing mean execution time
by 30.3\% (Table~\ref{tab:results}). Across the 169 paired tasks, \opensac answers 14 questions correctly that native
retrieval misses, while native retrieval answers nine that \opensac misses. Model token usage
falls by 37.8\%, accompanied by a 36.4\% reduction in recorded model cost.

On WideSearch, \opensac reduces mean execution time by 29.5\%, saving about
227 seconds per task, while row F1 declines by 0.84 percentage points.
Model token usage falls by 13.8\% and recorded model cost by 12.6\%.
The time reduction is similar across benchmarks, but token and model-cost
savings are larger on BrowseComp, where accuracy also improves.

\subsection{Case study: collecting labor statistics}
\label{sec:case-study}
The case study illustrates how the agent selects sources while code batches
retrieval and parses monthly records. In a recorded WideSearch run, the task
requests six monthly labor measures for
Alabama, Georgia, Louisiana, and Mississippi from January through June 2024.
Codex first searched for statistical series in a batch (Listing~\ref{lst:search}).
After examining the sources, it fetched Bureau of Labor Statistics (BLS) and
Federal Reserve Economic Data (FRED) endpoints in batches and parsed the responses
in Python (Listing~\ref{lst:fetch}). The complete run produced a 24-row Markdown
table. The excerpts omit query and URL setup, error checks, FRED CSV
parsing, and final formatting.

\begin{figure}[t]
\begin{lstlisting}[caption={Agent-written batch search for statistical sources
(excerpt).},label={lst:search}]
batches = sdk.search.many(queries, limit=8)
for q, result in zip(queries, batches, strict=True):
    print('\nQUERY:', q)
    if result.ok:
        for h in result.data:
            print(f'- {h.title} | {h.url}\n'
                  f'  {h.snippet[:400]}')
    else:
        print('ERROR', result.error.code, result.error.message)
\end{lstlisting}
\vspace{6pt}
\begin{lstlisting}[caption={Agent-written batch retrieval of BLS and FRED data
and BLS record parsing (excerpt).},label={lst:fetch}]
api_results = sdk.content.fetch_many(api_urls)
fred_results = sdk.content.fetch_many(fred_urls)
api_data = {}
for url, res in zip(api_urls, api_results, strict=True):
    sid = url.split('/timeseries/data/')[1].split('?')[0]
    fence = chr(96) * 3
    raw = re.sub(rf'^{fence}(?:json)?\s*|\s*{fence}\s*$',
                 '', res.data.text.strip(), flags=re.S)
    obj = json.loads(raw)
    dat = obj['Results']['series'][0]['data']
    api_data[sid] = {
        f"{x['year']}-{int(x['period'][1:]):02d}": x['value']
        for x in dat if x['period'].startswith('M')
        and x['period'] != 'M13'
    }
\end{lstlisting}
\end{figure}

\section{Conclusion and Future Work}
\opensac enables coding agents to compose retrieval operations in inspectable
programs and retain evidence for reuse. With skill guidance, it reduced
execution time and model token use on both benchmarks; accuracy improved on
BrowseComp, while row F1 declined slightly on WideSearch.

Improving how agents use SaC-like paradigms remains open. Better provider services could
expand source coverage and improve the relevance and reliability of evidence.
Skill guidance could be optimized through evaluation, including methods that
let agents propose and test revisions. Supervised fine-tuning and reinforcement
learning could help agents choose and compose operations and recover from
failures. OpenSaC's common interfaces, interchangeable providers, and
inspectable programs provide a foundation for future research.
\bibliographystyle{ACM-Reference-Format}
\bibliography{references}
\end{document}